\documentclass[11pt,a4paper]{article}
\usepackage[hyperref]{ranlp2021}
\usepackage{times}
\usepackage{latexsym}

\usepackage{microtype}
\usepackage{graphicx}

\aclfinalcopy 
\def\aclpaperid{94} 

\title{A Psychologically Informed Part-of-Speech Analysis of Depression in Social Media}

\author{Ana-Maria Bucur\textsuperscript{1} \\\And
  Ioana R. Podină\textsuperscript{2} \\
  \textsuperscript{1} Interdisciplinary School of Doctoral Studies \\
  \textsuperscript{2} Laboratory of Cognitive Clinical Sciences, Department of Psychology \\
  \textsuperscript{3} Faculty of Mathematics and Computer Science \\
  University of Bucharest, Romania\\
  \texttt{ana-maria.bucur@drd.unibuc.ro, ioana.podina@fpse.unibuc.ro} \\
  \texttt{ldinu@fmi.unibuc.ro} \\\And
  Liviu P. Dinu\textsuperscript{3} \\}

\date{}

\begin{document}
\maketitle
\begin{abstract}

In this work, we provide an extensive part-of-speech analysis of the discourse of social media users with depression. Research in psychology revealed that depressed users tend to be self-focused, more preoccupied with themselves and ruminate more about their lives and emotions. Our work aims to make use of large-scale datasets and computational methods for a quantitative exploration of discourse. We use the publicly available depression dataset from the Early Risk Prediction on the Internet Workshop (eRisk) 2018 and extract part-of-speech features and several indices based on them. Our results reveal statistically significant differences between the depressed and non-depressed individuals confirming findings from the existing psychology literature. Our work provides insights regarding the way in which depressed individuals are expressing themselves on social media platforms, allowing for better-informed computational models to help monitor and prevent mental illnesses. 

\end{abstract}

\section{Introduction}

Mental health disorders are a common problem in our world. Currently, mental health issues are on the rise: there is a 13\% increase in the past decade according to World Health Organization (WHO)\footnote{\url{https://www.who.int}}, with depression being at the forefront. Many mental illnesses remain undiagnosed due to social stigma, leading people to live 1 in 5 years of disability in their lifetime. 

With the rise of social media websites, interdisciplinary researchers in natural language processing, psychology and network analysis have turned their attention to automatically detect and monitor mental health manifestations through users' individual activity on social platforms (e.g. Facebook, Twitter, Reddit). The research is primarily focused on analyzing users' texts from posts and comments and determining, through computational linguistics models, the risk for various mental conditions - self-harm, depression, addictions etc.

Research is fulled through curated datasets \cite{yates2017depression, losada-crestani2016, amir-etal-2019-mental} with texts from primarily Reddit and Twitter. At the forefront of incentivising interdisciplinary research on monitoring mental health on social media are workshops such as the Early risk prediction on the Internet (eRisk) Workshop\footnote{\url{https://erisk.irlab.org/}} and the Workshop on Computational Linguistics and Clinical Psychology (CLPsych)\footnote{\url{https://clpsych.org/}}.

CLPsych and eRisk are two significant initiatives focusing on the interdisciplinary research area between computational linguistics and psychology. The eRisk Workshop, from the Conference and Labs of the Evaluation Forum (CLEF), focuses on the technologies that can be used for early risk detection of different pathologies or safety threats \cite{losadaoverview}. In five years of existence, the workshop addressed multiple mental health problems: pathological gambling, depression, self-harm and anorexia. 

The CLPsych Workshop was co-located with several international conferences on natural languages processing, the last edition was co-located with the Annual Conference of the North American Chapter of the Association for Computational Linguistics (NAACL). Throughout the seven editions of this workshop, it hosted shared tasks on depression, post-traumatic stress disorder (PTSD) \cite{coppersmith2015clpsych},  labeling crisis posts from the peer-support forum ReachOut\footnote{\url{https://au.reachout.com/}} \cite{milne2016clpsych}, predicting current and future psychological health from childhood essays \cite{lynn2018clpsych}, the degree of suicide risk  (no risk, low, moderate, or severe risk) \cite{zirikly2019clpsych} and suicide risk prediction from real data donated through OurDataHelps\footnote{\url{https://ourdatahelps.org/}} \cite{clpsych-2021-linguistics}.

In the present study, we perform a part-of-speech analysis and contribute to the understanding of the differences in social media discourse between depressed and non-depressed individuals. We focus on the differences in part-of-speech use and ground them in the existing literature from psychology researchers. We aim to answer the following two research questions:

\paragraph{RQ1:} Are there significant differences in part-of-speech use between individuals with self-reported depression diagnosis and control? 


\paragraph{RQ2:} Can these part-of-speech features be used alone to differentiate individuals with depression from controls? 


\section{Related Work}

Detecting the manifestations of mental health disorders from social media is an interdisciplinary research problem for researchers from psychology, natural language processing and network analysis. The two main approaches used to detect cues of depression from online users are: extracting linguistic features for a quantitative analysis or using automated models for classification.

The differences in language between depressed and non-depressed individuals focus on greater use of negative words, the personal pronoun \textit{"I"} \cite{rude2004language}, more words with negative polarity, frequent dichotomous expressions (e.g. \textit{always, never}) \cite{doi:10.1080/13811110214533}, cues of rumination (reflected in greater use of past tense verbs) \cite{smirnova2018language} in texts from depressed individuals.

Most linguistic features are extracted using the Linguistic inquiry and word count (LIWC) lexicon \cite{pennebaker2001linguistic}. LIWC provide a list of dictionary words for more than seventy categories: part-of-speech (e.g. personal pronouns, first-person personal pronouns, nouns, present/future/past verbs, adjectives), psychological processes (e.g. social, affective, cognitive processes), personal concerns (e.g. work, money, religion, death), etc. It is used to detect cues of depression \cite{loveys-etal-2018-cross, nalabandian, Eichstaedt11203}, neuroticism \cite{resnik-etal-2013-using}, to explore the language of suicide poets \cite{stirman2001word}, etc.

Other approaches to mental illnesses detection from text rely on character and word n-grams \cite{coppersmith-etal-2015-adhd, pedersen-2015-screening} or topic modelling techniques \cite{preotiuc-pietro-etal-2015-role, bucur2020detecting}.

Recent studies analyzing the online discourse of social media users with depression have focused on other particularities of language, such as offensive language. \citet{birnbaum2020identifying} found that depressed individuals use more swear words in their Facebook messages compared to controls. \citet{bucur-etal-2021-exploratory} apply offensive language identification techniques and show that users with depression diagnosis use more offensive language in their Reddit posts, individuals manifesting signs of depression in their posts having a more profane language and fewer insults targeted towards other individuals or groups.


Computational models used to detect the cues of depression from social media texts rely on traditional machine learning classifiers (e.g. SVM, Naïve Bayes) \cite{de2013social, aldarwish2017predicting, tadesse2019detection}, CNNs \cite{orabi2018deep, yates2017depression}, RNNs \cite{orabi2018deep} or transformer models \cite{martinez2020early, uban2020deep, bucur2021early}.

In the multimodal framework (involving text, voice and visual cues), the use of syntactic features (e.g. pronouns, adverbs usage) seems to improve the performance in depression detection, further emphasizing the relationship between linguistic features and depression \cite{morales-etal-2018-linguistically}.

Recently, researchers began exploring the interpretability and explainability of the decisions made by automatic classifiers for mental illnesses detection from social media to further understand the manifestations of different mental disorders in written language \cite{uban2021explainability, uban2021emotion}.

\section{Data}

In our experiments, we use the dataset from the eRisk workshop containing posts written in English from the social media platform Reddit.

The eRisk 2018 dataset \cite{losada-crestani2016} was created for the early detection of depression task. It contains two classes of users, depression and control. Users from the depression class were annotated by their mention of diagnosis in their posts (e.g. \textit{"I was diagnosed with depression"}), but expressions such as \textit{"I have depression", "I am depressed"} were not taken into account. The authors removed the mentions of diagnosis from the dataset. Users from the control group are random individuals who do not have any mention of diagnosis in their post, including those active in the depression subreddit\footnote{\url{https://www.reddit.com/r/depression/}}. The training dataset provided by the organizers contains 135 depressed users and 752 controls, while the test dataset contains 79 depressed users and 741 controls. We use both train and test splits in our exploration, consisting of a total of approximately 90,000 submissions from users with a depression diagnosis and over 985,000 in the control group.


\section{Methods}

\paragraph{Part-of-Speech Analysis} For each post in our dataset, we use the spaCy\footnote{\url{https://spacy.io/}} part-of-speech tagger to extract the corresponding tags and also morphological features (e.g. person and number for pronouns) for each word. We extract the universal POS tags\footnote{\url{https://universaldependencies.org/docs/u/pos/}} and the ones from The Penn Treebank tagset \cite{taylor2003penn}. We use the latter to explore the differences in the verb tenses. We assign the tenses according to their tags: VBD and VBN corresponding to present tense, VBG, VBZ and VBP corresponding to present tense, and MD tag before a VB corresponds to the future tense \cite{caragea2018exploring}. From the morphological features provided by the spaCy tagger, we extract the person and the number for all the pronouns. After this analysis, we use the following features in our exploration:
\paragraph{Universal Part-of-Speech:} ADJ, ADV, NOUN, PROPN, VERB, ADP, CCONJ, DET, PART, SCONJ, AUX, PRON
\paragraph{Verb tenses:} Past, present, future
\paragraph{Person of pronouns:} First, second and third-person
\paragraph{Pronoun number:} Singular and plural, only for the first-person

For each of these features, we compute their frequency for each post in the dataset. For the universal POS, the frequency is computed as the number of occurrences of a specific tag normalized by the total number of tags in a post. For verb tenses, the frequency of each tense is calculated as a percentage of the total number of verb occurrences. For the three kinds of personal pronouns, each frequency is computed as a percentage of the number of all personal pronouns. The frequency of singular and plural first-person pronouns is computed as a percentage of all first-person pronouns.

To further explore the part-of-speech usage in the social media dataset, we also use some special measurements \cite{havigerova2019text}:


\paragraph{Pronominalisation Index (PI):} reflecting the usage of pronouns, instead of another part-of-speech (e.g. nouns). It is computed as the number of pronouns divided by the number of nouns \cite{litvinova2016profiling}.

\paragraph{Formality Metric} \cite{mairesse2007using}:\vspace{2mm}
\vspace{2mm}
\hspace*{-0.1cm}
\scalebox{0.93}{%
$ {\scriptstyle F = \frac{NOUN + ADJ + PREP + ART - PRON - VERB - ADV - INTJ + 100}{2}}$}







Moreover, we test the discriminatory power of both POS tags frequency usage in users' texts and the specific computed indices. For this, we employ a Random Forest classifier on the train set of the eRisk 2018 dataset on the aforementioned features. To interpret the trained model and to estimate the importance of each feature, we employ SHapley Additive exPlanations (SHAP) \cite{NIPS2017_8a20a862} to measure each feature's contribution to the classifier decision. SHAP offers a game-theoretic approach to quantify feature importance, aligned with human intuition.

\paragraph{Classification} We opted for a simple Random Forest model, trained with 50 estimators and a max depth of 15, to avoid overfitting, with balanced class weights, since the dataset is heavily imbalanced. On the test set for the eRisk 2018 dataset, we obtain a weighted F$_1$-score of 78.37\% (with balanced class weights) and a macro F$_1$-score of 51.93\%. While the classification task is difficult, we are interested in exploring the feature importances of the model, which shed light on the model behaviour and provide us with insights regarding which POS tag is most discriminatory. 

We further present our findings and provide interpretations and discussions based on recent findings in psychology. 

\section{Results and Discussion}

Addressing our \textbf{RQ1}, we compute the Welch t-test for all our features and demonstrate that there are statistically significant differences ($p$-value $<$0.001) in part-of-speech usage between depressed and non-depressed individuals. In this section, we present these differences and their interpretation from the psychology research. 

\begin{figure}[!hbt]
    \centering
    \includegraphics[width=\linewidth]{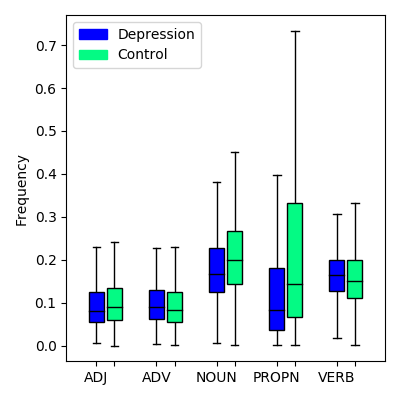}
    \caption{Frequency of content part-of-speech}
    \label{fig:content_pos_erisk2018}
\end{figure}

\paragraph{Content Part-of-speech} In Figure \ref{fig:content_pos_erisk2018}, we present the usage of content words for the two classes from the eRisk 2018 dataset. Individuals diagnosed with depression tend to use fewer common and proper nouns in comparison with control users. They also use more verbs and adverbs in their posts. The discourse is focused around actions, but with fewer entities (e.g. nouns), showing a defective linguistic structure with less interest in the environment (e.g. people, objects) \cite{de2016discovering}.

\begin{figure}[!hbt]
    \centering
    \includegraphics[width=1\linewidth]{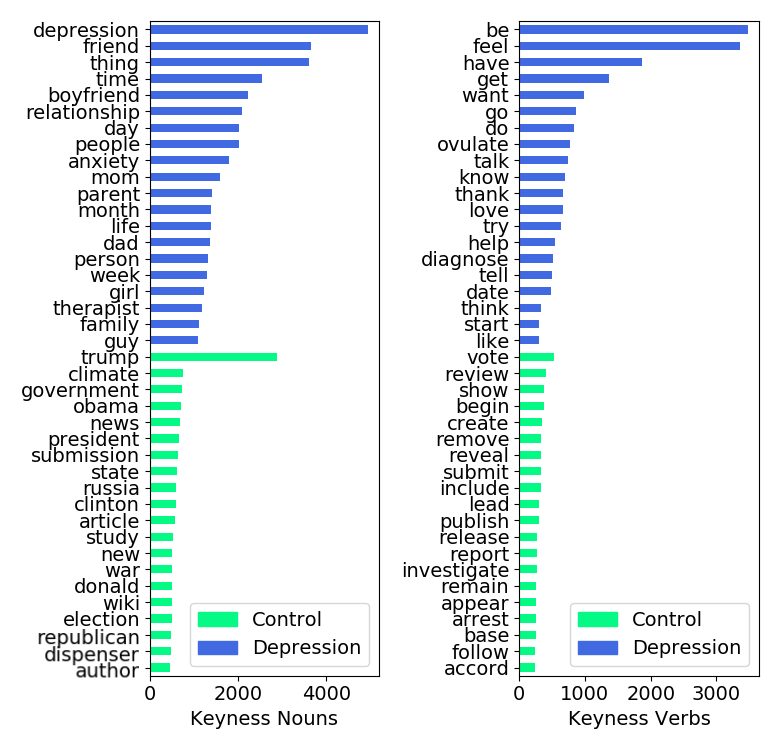}
    \caption{Most frequent nouns and verbs in each class}
    \label{fig:keyness_erisk2018}
\end{figure}

To further understand these differences, we pay a closer look at the frequencies of nouns and verbs in the social media discourse. We compute the keyness score \cite{kilgarriff2009simple,gabrielatos2018keyness} for verbs and nouns separately. In the keyness analysis, we compare the frequencies of nouns and verbs from the posts written by individuals with depression diagnosis (target corpus) in comparison to the posts from control users (reference corpus). In Figure \ref{fig:keyness_erisk2018}, we present the top 20 verbs and nouns from each corpus, ordered by their log-likelihood ratio (G$^2$) \cite{dunning1993accurate}.

\begin{figure}[!hbt]
    \centering
    \includegraphics[width=0.9\linewidth]{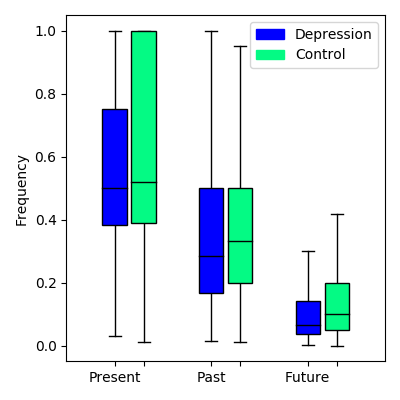}
    \caption{Frequency of verb tenses}
    \label{fig:verb_tenses_erisk2018}
\end{figure}

Rumination is a cognitive process focusing on past and present negative content and resulting in emotional distress \cite{sansone2012rumination}. It is present in several mental health disorders (e.g. depression, anxiety, obsessive-compulsive disorder, post-traumatic stress disorder). In depression, rumination is defined as \textit{behaviours and thoughts that focus one's attention on one's depressive symptoms and on the implications of these symptoms} \cite{nolen1991responses}. The rumination, as a response to depression, focuses the person's attention on their emotional state and inhibits the actions necessary to distract them from their mood. In Figure \ref{fig:verb_tenses_erisk2018},  when comparing the usage of the three verb tenses (present, past and future) between the two groups, we expected that signs of rumination would be present in our analysis of verb tenses, with texts from depressed users being shifted into the past \cite{smirnova2018language}, but this result is not found in this sample of individuals. 

Regarding the usage of future tense, depressed people have a lower frequency of verbs portraying future actions. This result may be a consequence of anhedonia, people suffering from depression reporting lower anticipatory pleasure, and thus talking online less about their future plans. Anhedonia, defined as \textit{markedly diminished interest or pleasure in all, or almost all activities most of the day, nearly every day} \cite{american2013diagnostic}, is a common symptom of depression.

The higher frequency of cognitive verbs (e.g. \textit{feel, think, know}) in the texts written by depressed individuals indicates the cognitive impairments and judgement issues specific to depression. People with depression have cognitive deficits and biases in the processing of emotional information and they are unable to adaptively regulate their emotions \cite{de2014mental}. Individuals with or without depression may not differ in their initial response to an adverse event. Still, they differ in their ability to recover once they have experienced the negative emotion. Depressed individuals are not able to repair their mood. Instead, they remain in a negative state of mind, which is related to increased negative thoughts, selective attention to negative stimuli and greater accessibility of negative recollections \cite{joormann2010cognitive}. In comparison, the individuals from the control group use more action verbs (e.g. \textit{vote, lead, show, begin, create}. In addition, depressed individuals are more passive; they have a lower level of general activity, consistent with symptoms of depressive disorder \cite{hopko2003use}.

Being an online social media platform similar to forums, Reddit is organized in subreddits with specific topics. It also has several communities dedicated to mental health problems. Compared to other social media platforms that require real-name authentication (e.g. Facebook), Reddit affords users anonymity or pseudo-anonymity. Complete anonymity is almost impossible, users providing bits of information with every interaction on the platform (e.g. comments, posts). Reddit allows users to create \textit{throwaway} accounts to engage temporary without revealing their identity \cite{kilgo2018reddit}. These types of accounts are used to discuss sensitive information or stigmatizing problems. \citet{de2014mental} study the mental health discourse on Reddit and show that its communities allow a high degree of information exchange related to mental health. Users use Reddit to self-disclose the challenges faced in their daily lives or in personal relationships. They also seek emotional support or specific information about mental illnesses diagnosis and treatment. Their study demonstrates that Reddit fills the gap between social media platforms like Twitter and Facebook and online health forums regarding mental health discourse. 

Examining the frequencies of nouns in the eRisk 2018 dataset, we show in Figure \ref{fig:keyness_erisk2018} that the users with self-reported depression diagnosis use their Reddit account to disclose and discuss their mental health problems (e.g. \textit{depression, anxiety, therapist}) or their personal relationships (e.g. \textit{friend, boyfriend, relationship, mom, dad}). The process of seeking online support is also shown in the frequency of verbs: \textit{feel, talk, diagnose, help}.

Even if the dataset contains control users active in the depression subreddit, the majority of control users seem to post on other themed subreddits (e.g. politics). \citet{bucur2020detecting} perform a topic modelling analysis and show that texts from control users are found in topics related to their hobbies, as opposed to depressed people, who are more focused on their feelings and life events. Our results are in line with this study, the users from the control group use more politics-related words (e.g. \textit{trump, government, president, news, vote}). 


\begin{figure}[!hbt]
    \centering
    \includegraphics[width=\linewidth]{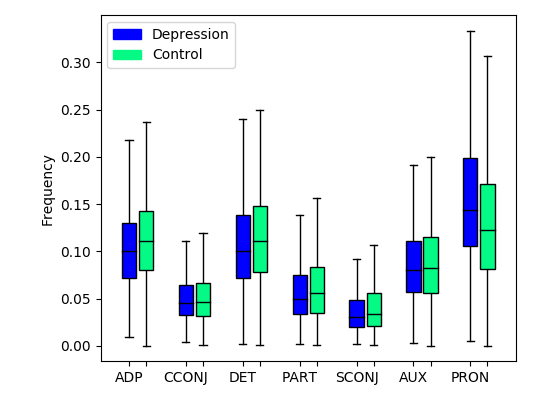}
    \caption{Frequency of functional part-of-speech}
    \label{fig:func_pos_erisk2018}
\end{figure}

\paragraph{Functional part-of-speech} In Figure \ref{fig:func_pos_erisk2018}, we present the frequencies of functional part-of-speech for depressed and control users. Depressed individuals generally use fewer functional words in their texts in comparison with control users, with the exception of pronouns. Neurons involved in content words processing are equally distributed over both hemispheres, while function words are processed mainly in the left hemisphere \cite{pulvermuller1995electrocortical}. Lower preposition usage may be an outcome of deficient activation of the left hemisphere regions, responsible for producing more abstract lexical units \cite{litvinova2016profiling}. Function words are highly social, the capacity to use function words requires social skills \cite{mindmapping}.

\begin{figure}[!hbt]
    \centering
    \includegraphics[width=0.9\linewidth]{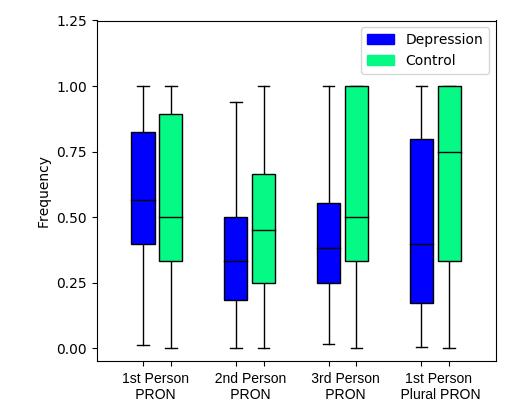}
    \caption{Frequency of personal pronouns}
    \label{fig:pron_erisk2018}
\end{figure}

The Figure \ref{fig:pron_erisk2018} shows the
differences in personal pronouns used by  the two groups. The high frequency of first-person singular pronouns indicates a higher self-preoccupation in depressed individuals, as opposed to the first-person plural pronouns, which shows collective attention. Second and third-person pronouns indicate social interactivity and contain references to other people or things in the environment \cite{de2016discovering}. 

Depressed users use more first-person singular pronouns. The frequencies of first-person plural pronouns are inversely proportional to the first-person singular pronouns frequencies. This result is in line with the self-focused attention tendency (SFA) in depressed individuals. SFA is a cognitive bias linked to depression; the high frequency of first-person singular pronouns in speech or written text is considered a linguistic marker of SFA \cite{brockmeyer2015me}. Individuals with depression have deficits in other-focused social cognitions, they are impaired in Theory of Mind reasoning and empathy. Theory of Mind enables people to make inferences on the behaviour of others and their own \cite{premack1978does}. \citet{erle2019egocentrism} show that individuals exhibiting high levels of depressive symptoms were impaired on tasks involving overcoming their egocentrism.  

The usage of fewer first-person plural pronouns in texts from users diagnosed with depression may be a sign of a lesser sense of belonging. The information-processing biases of depressed individuals make it hard for them to perceive cues of acceptance and belonging in social interactions, and to view ambiguous social interactions as being negative \cite{steger2009depression}. 

\begin{figure}[!hbt]
    \centering
    \includegraphics[width=0.8\linewidth]{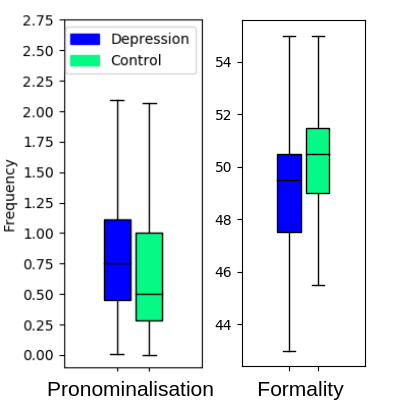}
    \caption{Indices based on part-of-speech tags}
    \label{fig:indexes}
\end{figure}

\begin{figure}[hbt!]
    \centering
    \includegraphics[width=1.0\linewidth]{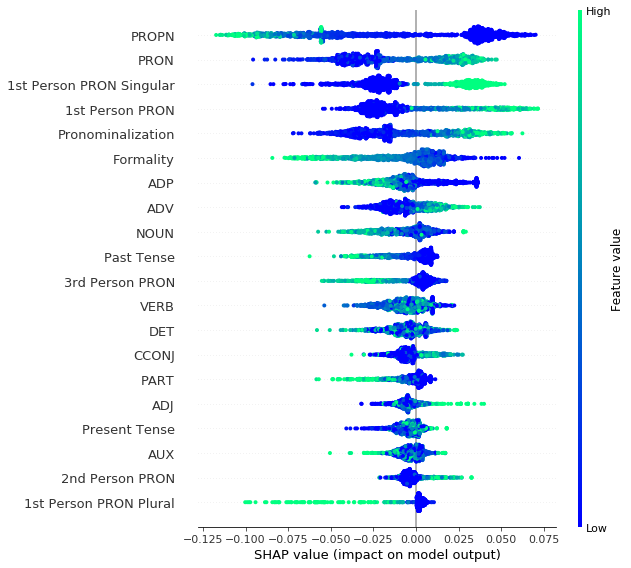}
    \caption{Feature importances of the RandomForest model for depression classification using POS frequencies and POS indices. As shown in our analysis, the usage of pronouns (1st Person Singular/Plural) and proper nouns is the most discriminatory.}
    \label{fig:shap-indices}
\end{figure}

\paragraph{Pronominalisation and Formality Indices}In Figure \ref{fig:indexes} we present the results for the two metrics computed on part-of-speech tags. 
Depressed individuals have a higher pronominalisation index, using more pronouns in spite of nouns. This finding is also found in the language of people with self-destructive behaviour, insufficient activation of the cerebellum being associated with suicidal behaviour \cite{litvinova2016profiling}. Depressive individuals also use less formal language. They have a more contextualized discourse on social media, providing information about the context in order to avoid ambiguity \cite{Contextuality}. 

Addressing our \textbf{RQ2}, we show in Figure \ref{fig:shap-indices} the Shapley values for a random sample of 1500 posts from the eRisk 2018 test set. A higher Shapley value corresponds with a higher importance in the final model decision based on the feature value. We used all computed POS features and indices in our model, but show only the top 20 for clarity. It is evident that from the summary plot, the absence of proper nouns is the most discriminatory factor in the decision to classify a person as depressive. Moreover, the use of pronouns (also evident in the Pronominalisation Index) is highly correlated with positive model output. The high usage of first-person singular pronouns and low usage of first-person plural pronouns confirms both our findings in the exploratory analysis and psychology literature.


\section{Conclusion}

In this work, we provide an extensive analysis of part-of-speech usage in social media texts from depressed and non-depressed users. Our findings are confirmed by studies in psychology and show that individuals diagnosed with depression use more pronouns (especially first-person singular pronouns) and verbs, and fewer common and proper nouns. Their social media discourse revolves around their life experiences and sentiments, as opposed to control users who are not interested in discussing their problems online.

Moreover, we also provided insights into the discriminatory power of POS frequencies by employing SHAP, a game-theoretic approach for model interpretation. Through this, we showed that depressive users can be characterized most easily, primarily through their usage of pronouns and proper nouns. 


\bibliographystyle{acl_natbib}
\bibliography{ranlp2021}

\begin{thebibliography}{58}
\expandafter\ifx\csname natexlab\endcsname\relax\def\natexlab#1{#1}\fi

\bibitem[{Aldarwish and Ahmad(2017)}]{aldarwish2017predicting}
Maryam~Mohammed Aldarwish and Hafiz~Farooq Ahmad. 2017.
\newblock Predicting depression levels using social media posts.
\newblock In \emph{2017 IEEE 13th international Symposium on Autonomous
  decentralized system (ISADS)}, pages 277--280. IEEE.

\bibitem[{Amir et~al.(2019)Amir, Dredze, and Ayers}]{amir-etal-2019-mental}
Silvio Amir, Mark Dredze, and John~W. Ayers. 2019.
\newblock \href {https://doi.org/10.18653/v1/W19-3013} {Mental health
  surveillance over social media with digital cohorts}.
\newblock In \emph{Proceedings of the Sixth Workshop on Computational
  Linguistics and Clinical Psychology}, pages 114--120, Minneapolis, Minnesota.
  Association for Computational Linguistics.

\bibitem[{Association et~al.(2013)}]{american2013diagnostic}
American~Psychiatric Association et~al. 2013.
\newblock \emph{Diagnostic and statistical manual of mental disorders
  (DSM-5{\textregistered})}.
\newblock American Psychiatric Pub.

\bibitem[{Birnbaum et~al.(2020)Birnbaum, Norel, Van~Meter, Ali, Arenare,
  Eyigoz, Agurto, Germano, Kane, and Cecchi}]{birnbaum2020identifying}
Michael~L Birnbaum, Raquel Norel, Anna Van~Meter, Asra~F Ali, Elizabeth
  Arenare, Elif Eyigoz, Carla Agurto, Nicole Germano, John~M Kane, and
  Guillermo~A Cecchi. 2020.
\newblock Identifying signals associated with psychiatric illness utilizing
  language and images posted to facebook.
\newblock \emph{npj Schizophrenia}, 6(1):1--10.

\bibitem[{Brockmeyer et~al.(2015)Brockmeyer, Zimmermann, Kulessa, Hautzinger,
  Bents, Friederich, Herzog, and Backenstrass}]{brockmeyer2015me}
Timo Brockmeyer, Johannes Zimmermann, Dominika Kulessa, Martin Hautzinger,
  Hinrich Bents, Hans-Christoph Friederich, Wolfgang Herzog, and Matthias
  Backenstrass. 2015.
\newblock Me, myself, and i: self-referent word use as an indicator of
  self-focused attention in relation to depression and anxiety.
\newblock \emph{Frontiers in psychology}, 6:1564.

\bibitem[{Bucur et~al.(2021{\natexlab{a}})Bucur, Cosma, and
  Dinu}]{bucur2021early}
Ana-Maria Bucur, Adrian Cosma, and Liviu~P Dinu. 2021{\natexlab{a}}.
\newblock Early risk detection of pathological gambling, self-harm and
  depression using bert.
\newblock \emph{CLEF (Working Notes)}.

\bibitem[{Bucur and Dinu(2020)}]{bucur2020detecting}
Ana-Maria Bucur and Liviu~P Dinu. 2020.
\newblock Detecting early onset of depression from social media text using
  learned confidence scores.
\newblock \emph{Proceedings of the Seventh Italian Conference on Computational
  Linguistics}.

\bibitem[{Bucur et~al.(2021{\natexlab{b}})Bucur, Zampieri, and
  Dinu}]{bucur-etal-2021-exploratory}
Ana-Maria Bucur, Marcos Zampieri, and Liviu~P. Dinu. 2021{\natexlab{b}}.
\newblock \href {https://doi.org/10.18653/v1/2021.findings-acl.315} {An
  exploratory analysis of the relation between offensive language and mental
  health}.
\newblock In \emph{Findings of the Association for Computational Linguistics:
  ACL-IJCNLP 2021}, pages 3600--3606, Online. Association for Computational
  Linguistics.

\bibitem[{Caragea et~al.(2018)Caragea, Dinu, and
  Dumitru}]{caragea2018exploring}
Cornelia Caragea, Liviu~P Dinu, and Bogdan Dumitru. 2018.
\newblock Exploring optimism and pessimism in twitter using deep learning.
\newblock In \emph{Proceedings of the 2018 Conference on Empirical Methods in
  Natural Language Processing}, pages 652--658.

\bibitem[{Coppersmith et~al.(2015{\natexlab{a}})Coppersmith, Dredze, Harman,
  and Hollingshead}]{coppersmith-etal-2015-adhd}
Glen Coppersmith, Mark Dredze, Craig Harman, and Kristy Hollingshead.
  2015{\natexlab{a}}.
\newblock \href {https://doi.org/10.3115/v1/W15-1201} {From {ADHD} to {SAD}:
  Analyzing the language of mental health on {T}witter through self-reported
  diagnoses}.
\newblock In \emph{Proceedings of the 2nd Workshop on Computational Linguistics
  and Clinical Psychology: From Linguistic Signal to Clinical Reality}, pages
  1--10, Denver, Colorado. Association for Computational Linguistics.

\bibitem[{Coppersmith et~al.(2015{\natexlab{b}})Coppersmith, Dredze, Harman,
  Hollingshead, and Mitchell}]{coppersmith2015clpsych}
Glen Coppersmith, Mark Dredze, Craig Harman, Kristy Hollingshead, and Margaret
  Mitchell. 2015{\natexlab{b}}.
\newblock Clpsych 2015 shared task: Depression and ptsd on twitter.
\newblock In \emph{Proceedings of the 2nd Workshop on Computational Linguistics
  and Clinical Psychology: From Linguistic Signal to Clinical Reality}, pages
  31--39.

\bibitem[{De~Choudhury et~al.(2013)De~Choudhury, Counts, and
  Horvitz}]{de2013social}
Munmun De~Choudhury, Scott Counts, and Eric Horvitz. 2013.
\newblock Social media as a measurement tool of depression in populations.
\newblock In \emph{Proceedings of the 5th annual ACM web science conference},
  pages 47--56.

\bibitem[{De~Choudhury and De(2014)}]{de2014mental}
Munmun De~Choudhury and Sushovan De. 2014.
\newblock Mental health discourse on reddit: Self-disclosure, social support,
  and anonymity.
\newblock In \emph{Proceedings of the International AAAI Conference on Web and
  Social Media}, volume~8.

\bibitem[{De~Choudhury et~al.(2016)De~Choudhury, Kiciman, Dredze, Coppersmith,
  and Kumar}]{de2016discovering}
Munmun De~Choudhury, Emre Kiciman, Mark Dredze, Glen Coppersmith, and Mrinal
  Kumar. 2016.
\newblock Discovering shifts to suicidal ideation from mental health content in
  social media.
\newblock In \emph{Proceedings of the 2016 CHI conference on human factors in
  computing systems}, pages 2098--2110.

\bibitem[{Dunning(1993)}]{dunning1993accurate}
Ted~E Dunning. 1993.
\newblock Accurate methods for the statistics of surprise and coincidence.
\newblock \emph{Computational linguistics}, 19(1):61--74.

\bibitem[{Eichstaedt et~al.(2018)Eichstaedt, Smith, Merchant, Ungar, Crutchley,
  Preo{\c t}iuc-Pietro, Asch, and Schwartz}]{Eichstaedt11203}
Johannes~C. Eichstaedt, Robert~J. Smith, Raina~M. Merchant, Lyle~H. Ungar,
  Patrick Crutchley, Daniel Preo{\c t}iuc-Pietro, David~A. Asch, and H.~Andrew
  Schwartz. 2018.
\newblock \href {https://doi.org/10.1073/pnas.1802331115} {Facebook language
  predicts depression in medical records}.
\newblock \emph{Proceedings of the National Academy of Sciences},
  115(44):11203--11208.

\bibitem[{Erle et~al.(2019)Erle, Barth, and Topolinski}]{erle2019egocentrism}
Thorsten~M Erle, Niklas Barth, and Sascha Topolinski. 2019.
\newblock Egocentrism in sub-clinical depression.
\newblock \emph{Cognition and Emotion}, 33(6):1239--1248.

\bibitem[{Fekete(2002)}]{doi:10.1080/13811110214533}
Sandor Fekete. 2002.
\newblock \href {https://doi.org/10.1080/13811110214533} {The internet - a new
  source of data on suicide, depression and anxiety: A preliminary study}.
\newblock \emph{Archives of Suicide Research}, 6(4):351--361.

\bibitem[{Gabrielatos(2018)}]{gabrielatos2018keyness}
Costas Gabrielatos. 2018.
\newblock Keyness analysis.
\newblock \emph{Corpus approaches to discourse: A critical review}, pages
  225--258.

\bibitem[{Goharian et~al.(2021)Goharian, Resnik, Yates, Ireland, Niederhoffer,
  and Resnik}]{clpsych-2021-linguistics}
Nazli Goharian, Philip Resnik, Andrew Yates, Molly Ireland, Kate Niederhoffer,
  and Rebecca Resnik, editors. 2021.
\newblock \href {https://www.aclweb.org/anthology/2021.clpsych-1.0}
  {\emph{Proceedings of the Seventh Workshop on Computational Linguistics and
  Clinical Psychology: Improving Access}}. Association for Computational
  Linguistics, Online.

\bibitem[{Havigerov{\'a} et~al.(2019)Havigerov{\'a}, Haviger, Ku{\v{c}}era, and
  Hoffmannov{\'a}}]{havigerova2019text}
Jana~M Havigerov{\'a}, Ji{\v{r}}{\'\i} Haviger, Dalibor Ku{\v{c}}era, and Petra
  Hoffmannov{\'a}. 2019.
\newblock Text-based detection of the risk of depression.
\newblock \emph{Frontiers in psychology}, 10:513.

\bibitem[{Heylighen and Dewaele(2002)}]{Contextuality}
Francis Heylighen and Jean-Marc Dewaele. 2002.
\newblock \href {https://doi.org/10.1023/A:1019661126744} {Variation in the
  contextuality of language: An empirical measure}.
\newblock \emph{Foundations of Science}, 7:293--340.

\bibitem[{Hopko et~al.(2003)Hopko, Armento, Cantu, Chambers, and
  Lejuez}]{hopko2003use}
Derek~R Hopko, Maria~EA Armento, Melissa~S Cantu, Laura~L Chambers, and
  CW~Lejuez. 2003.
\newblock The use of daily diaries to assess the relations among mood state,
  overt behavior, and reward value of activities.
\newblock \emph{Behaviour research and therapy}, 41(10):1137--1148.

\bibitem[{Joormann(2010)}]{joormann2010cognitive}
Jutta Joormann. 2010.
\newblock Cognitive inhibition and emotion regulation in depression.
\newblock \emph{Current Directions in Psychological Science}, 19(3):161--166.

\bibitem[{Kilgarriff(2009)}]{kilgarriff2009simple}
Adam Kilgarriff. 2009.
\newblock Simple maths for keywords.
\newblock In \emph{Proceedings of the CL}.

\bibitem[{Kilgo et~al.(2018)Kilgo, Ng, Riedl, and Lacasa-Mas}]{kilgo2018reddit}
Danielle~K Kilgo, Yee Man~Margaret Ng, Martin~J Riedl, and Ivan Lacasa-Mas.
  2018.
\newblock Reddit’s veil of anonymity: Predictors of engagement and
  participation in media environments with hostile reputations.
\newblock \emph{Social Media+ Society}, 4(4):2056305118810216.

\bibitem[{Litvinova et~al.(2016)Litvinova, Zagorovskaya, Litvinova, and
  Seredin}]{litvinova2016profiling}
Tatiana Litvinova, Olga Zagorovskaya, Olga Litvinova, and Pavel Seredin. 2016.
\newblock Profiling a set of personality traits of a text’s author: a
  corpus-based approach.
\newblock In \emph{International Conference on Speech and Computer}, pages
  555--562. Springer.

\bibitem[{Losada and Crestani(2016)}]{losada-crestani2016}
D.~Losada and F.~Crestani. 2016.
\newblock A test collection for research on depression and language use.
\newblock In \emph{Proc. of Experimental IR Meets Multilinguality,
  Multimodality, and Interaction, 7th International Conference of the {CLEF}
  Association, {CLEF} 2016}, pages 28--39, Evora, Portugal.

\bibitem[{Losada et~al.()Losada, Crestani, and Parapar}]{losadaoverview}
David~E Losada, Fabio Crestani, and Javier Parapar.
\newblock Overview of erisk at clef 2020: Early risk prediction on the internet
  (extended overview).

\bibitem[{Loveys et~al.(2018)Loveys, Torrez, Fine, Moriarty, and
  Coppersmith}]{loveys-etal-2018-cross}
Kate Loveys, Jonathan Torrez, Alex Fine, Glen Moriarty, and Glen Coppersmith.
  2018.
\newblock \href {https://doi.org/10.18653/v1/W18-0608} {Cross-cultural
  differences in language markers of depression online}.
\newblock In \emph{Proceedings of the Fifth Workshop on Computational
  Linguistics and Clinical Psychology: From Keyboard to Clinic}, pages 78--87,
  New Orleans, LA. Association for Computational Linguistics.

\bibitem[{Lundberg and Lee(2017)}]{NIPS2017_8a20a862}
Scott~M Lundberg and Su-In Lee. 2017.
\newblock \href
  {https://proceedings.neurips.cc/paper/2017/file/8a20a8621978632d76c43dfd28b67767-Paper.pdf}
  {A unified approach to interpreting model predictions}.
\newblock In \emph{Advances in Neural Information Processing Systems},
  volume~30. Curran Associates, Inc.

\bibitem[{Lynn et~al.(2018)Lynn, Goodman, Niederhoffer, Loveys, Resnik, and
  Schwartz}]{lynn2018clpsych}
Veronica Lynn, Alissa Goodman, Kate Niederhoffer, Kate Loveys, Philip Resnik,
  and H~Andrew Schwartz. 2018.
\newblock Clpsych 2018 shared task: Predicting current and future psychological
  health from childhood essays.
\newblock In \emph{Proceedings of the Fifth Workshop on Computational
  Linguistics and Clinical Psychology: From Keyboard to Clinic}, pages 37--46.

\bibitem[{Mairesse et~al.(2007)Mairesse, Walker, Mehl, and
  Moore}]{mairesse2007using}
Fran{\c{c}}ois Mairesse, Marilyn~A Walker, Matthias~R Mehl, and Roger~K Moore.
  2007.
\newblock Using linguistic cues for the automatic recognition of personality in
  conversation and text.
\newblock \emph{Journal of artificial intelligence research}, 30:457--500.

\bibitem[{Mart{\'\i}nez-Casta{\~n}o et~al.(2020)Mart{\'\i}nez-Casta{\~n}o,
  Htait, Azzopardi, and Moshfeghi}]{martinez2020early}
Rodrigo Mart{\'\i}nez-Casta{\~n}o, Amal Htait, Leif Azzopardi, and Yashar
  Moshfeghi. 2020.
\newblock Early risk detection of self-harm and depression severity using
  bert-based transformers: ilab at clef erisk 2020.
\newblock \emph{Early Risk Prediction on the Internet}.

\bibitem[{Milne et~al.(2016)Milne, Pink, Hachey, and Calvo}]{milne2016clpsych}
David~N Milne, Glen Pink, Ben Hachey, and Rafael~A Calvo. 2016.
\newblock Clpsych 2016 shared task: Triaging content in online peer-support
  forums.
\newblock In \emph{Proceedings of the third workshop on computational
  linguistics and clinical psychology}, pages 118--127.

\bibitem[{Morales et~al.(2018)Morales, Scherer, and
  Levitan}]{morales-etal-2018-linguistically}
Michelle Morales, Stefan Scherer, and Rivka Levitan. 2018.
\newblock \href {https://doi.org/10.18653/v1/W18-0602} {A
  linguistically-informed fusion approach for multimodal depression detection}.
\newblock In \emph{Proceedings of the Fifth Workshop on Computational
  Linguistics and Clinical Psychology: From Keyboard to Clinic}, pages 13--24,
  New Orleans, LA. Association for Computational Linguistics.

\bibitem[{Nalabandian and Ireland(2019)}]{nalabandian}
Taleen Nalabandian and Molly Ireland. 2019.
\newblock \href {https://doi.org/10.18653/v1/W19-3008} {Depressed individuals
  use negative self-focused language when recalling recent interactions with
  close romantic partners but not family or friends}.
\newblock pages 62--73.

\bibitem[{Nolen-Hoeksema(1991)}]{nolen1991responses}
Susan Nolen-Hoeksema. 1991.
\newblock Responses to depression and their effects on the duration of
  depressive episodes.
\newblock \emph{Journal of abnormal psychology}, 100(4):569.

\bibitem[{Orabi et~al.(2018)Orabi, Buddhitha, Orabi, and
  Inkpen}]{orabi2018deep}
Ahmed~Husseini Orabi, Prasadith Buddhitha, Mahmoud~Husseini Orabi, and Diana
  Inkpen. 2018.
\newblock Deep learning for depression detection of twitter users.
\newblock In \emph{Proceedings of the Fifth Workshop on Computational
  Linguistics and Clinical Psychology: From Keyboard to Clinic}, pages 88--97.

\bibitem[{Pedersen(2015)}]{pedersen-2015-screening}
Ted Pedersen. 2015.
\newblock \href {https://doi.org/10.3115/v1/W15-1206} {Screening {T}witter
  users for depression and {PTSD} with lexical decision lists}.
\newblock In \emph{Proceedings of the 2nd Workshop on Computational Linguistics
  and Clinical Psychology: From Linguistic Signal to Clinical Reality}, pages
  46--53, Denver, Colorado. Association for Computational Linguistics.

\bibitem[{Pennebaker(2017)}]{mindmapping}
James Pennebaker. 2017.
\newblock \href {https://doi.org/10.1016/j.procs.2017.11.150} {Mind mapping:
  Using everyday language to explore social \& psychological processes}.
\newblock \emph{Procedia Computer Science}, 118:100--107.

\bibitem[{Pennebaker et~al.(2001)Pennebaker, Francis, and
  Booth}]{pennebaker2001linguistic}
James~W Pennebaker, Martha~E Francis, and Roger~J Booth. 2001.
\newblock Linguistic inquiry and word count: Liwc 2001.
\newblock \emph{Mahway: Lawrence Erlbaum Associates}, 71(2001):2001.

\bibitem[{Premack and Woodruff(1978)}]{premack1978does}
David Premack and Guy Woodruff. 1978.
\newblock Does the chimpanzee have a theory of mind?
\newblock \emph{Behavioral and brain sciences}, 1(4):515--526.

\bibitem[{Preo{\c{t}}iuc-Pietro et~al.(2015)Preo{\c{t}}iuc-Pietro, Eichstaedt,
  Park, Sap, Smith, Tobolsky, Schwartz, and
  Ungar}]{preotiuc-pietro-etal-2015-role}
Daniel Preo{\c{t}}iuc-Pietro, Johannes Eichstaedt, Gregory Park, Maarten Sap,
  Laura Smith, Victoria Tobolsky, H.~Andrew Schwartz, and Lyle Ungar. 2015.
\newblock \href {https://doi.org/10.3115/v1/W15-1203} {The role of personality,
  age, and gender in tweeting about mental illness}.
\newblock In \emph{Proceedings of the 2nd Workshop on Computational Linguistics
  and Clinical Psychology: From Linguistic Signal to Clinical Reality}, pages
  21--30, Denver, Colorado. Association for Computational Linguistics.

\bibitem[{Pulverm{\"u}ller et~al.(1995)Pulverm{\"u}ller, Lutzenberger, and
  Birbaumer}]{pulvermuller1995electrocortical}
Friedemann Pulverm{\"u}ller, Werner Lutzenberger, and Niels Birbaumer. 1995.
\newblock Electrocortical distinction of vocabulary types.
\newblock \emph{Electroencephalography and clinical Neurophysiology},
  94(5):357--370.

\bibitem[{Resnik et~al.(2013)Resnik, Garron, and
  Resnik}]{resnik-etal-2013-using}
Philip Resnik, Anderson Garron, and Rebecca Resnik. 2013.
\newblock \href {https://www.aclweb.org/anthology/D13-1133} {Using topic
  modeling to improve prediction of neuroticism and depression in college
  students}.
\newblock In \emph{Proceedings of the 2013 Conference on Empirical Methods in
  Natural Language Processing}, pages 1348--1353, Seattle, Washington, USA.
  Association for Computational Linguistics.

\bibitem[{Rude et~al.(2004)Rude, Gortner, and Pennebaker}]{rude2004language}
Stephanie Rude, Eva-Maria Gortner, and James Pennebaker. 2004.
\newblock Language use of depressed and depression-vulnerable college students.
\newblock \emph{Cognition \& Emotion}, 18(8):1121--1133.

\bibitem[{Sansone and Sansone(2012)}]{sansone2012rumination}
Randy~A Sansone and Lori~A Sansone. 2012.
\newblock Rumination: relationships with physical health.
\newblock \emph{Innovations in clinical neuroscience}, 9(2):29.

\bibitem[{Smirnova et~al.(2018)Smirnova, Cumming, Sloeva, Kuvshinova, Romanov,
  and Nosachev}]{smirnova2018language}
Daria Smirnova, Paul Cumming, Elena Sloeva, Natalia Kuvshinova, Dmitry Romanov,
  and Gennadii Nosachev. 2018.
\newblock Language patterns discriminate mild depression from normal sadness
  and euthymic state.
\newblock \emph{Frontiers in psychiatry}, 9:105.

\bibitem[{Steger and Kashdan(2009)}]{steger2009depression}
Michael~F Steger and Todd~B Kashdan. 2009.
\newblock Depression and everyday social activity, belonging, and well-being.
\newblock \emph{Journal of counseling psychology}, 56(2):289.

\bibitem[{Stirman and Pennebaker(2001)}]{stirman2001word}
Shannon~Wiltsey Stirman and James~W Pennebaker. 2001.
\newblock Word use in the poetry of suicidal and nonsuicidal poets.
\newblock \emph{Psychosomatic Medicine}, 63(4):517--522.

\bibitem[{Tadesse et~al.(2019)Tadesse, Lin, Xu, and
  Yang}]{tadesse2019detection}
Michael~M Tadesse, Hongfei Lin, Bo~Xu, and Liang Yang. 2019.
\newblock Detection of depression-related posts in reddit social media forum.
\newblock \emph{IEEE Access}, 7:44883--44893.

\bibitem[{Taylor et~al.(2003)Taylor, Marcus, and Santorini}]{taylor2003penn}
Ann Taylor, Mitchell Marcus, and Beatrice Santorini. 2003.
\newblock The penn treebank: an overview.
\newblock \emph{Treebanks}, pages 5--22.

\bibitem[{Uban et~al.(2021{\natexlab{a}})Uban, Chulvi, and
  Rosso}]{uban2021emotion}
Ana-Sabina Uban, Berta Chulvi, and Paolo Rosso. 2021{\natexlab{a}}.
\newblock An emotion and cognitive based analysis of mental health disorders
  from social media data.
\newblock \emph{Future Generation Computer Systems}.

\bibitem[{Uban et~al.(2021{\natexlab{b}})Uban, Chulvi, and
  Rosso}]{uban2021explainability}
Ana~Sabina Uban, Berta Chulvi, and Paolo Rosso. 2021{\natexlab{b}}.
\newblock On the explainability of automatic predictions of mental disorders
  from social media data.
\newblock In \emph{International Conference on Applications of Natural Language
  to Information Systems}, pages 301--314. Springer.

\bibitem[{Uban and Rosso(2020)}]{uban2020deep}
Ana-Sabina Uban and Paolo Rosso. 2020.
\newblock Deep learning architectures and strategies for early detection of
  self-harm and depression level prediction.
\newblock In \emph{CEUR Workshop Proceedings}, volume 2696, pages 1--12. Sun
  SITE Central Europe.

\bibitem[{Yates et~al.(2017)Yates, Cohan, and Goharian}]{yates2017depression}
Andrew Yates, Arman Cohan, and Nazli Goharian. 2017.
\newblock Depression and self-harm risk assessment in online forums.
\newblock In \emph{Proceedings of the 2017 Conference on Empirical Methods in
  Natural Language Processing}, pages 2968--2978.

\bibitem[{Zirikly et~al.(2019)Zirikly, Resnik, Uzuner, and
  Hollingshead}]{zirikly2019clpsych}
Ayah Zirikly, Philip Resnik, Ozlem Uzuner, and Kristy Hollingshead. 2019.
\newblock Clpsych 2019 shared task: Predicting the degree of suicide risk in
  reddit posts.
\newblock In \emph{Proceedings of the sixth workshop on computational
  linguistics and clinical psychology}, pages 24--33.

\end{thebibliography}


\end{document}